\documentclass{article}
\usepackage[preprint]{spconf}
\usepackage{amsmath,graphicx}
\copyrightnotice{\copyright\ IEEE 2023}
\toappear{To appear in {\it Proc.\ ICASSP2023,
                   June 04-10, 2023, Rhodes Island, Greece}}

\usepackage{booktabs}
\usepackage{cite,url}

\title{End-to-end Spoken Language Understanding \\with Tree-constrained Pointer Generator}
\name{Guangzhi Sun, Chao Zhang, Philip C. Woodland
\thanks{Guangzhi Sun is supported by a Cambridge International Scholarship from the Cambridge Trust. This work has been performed using resources provided by the Cambridge Tier-2 system operated by the University of Cambridge Research Computing Service (www.hpc.cam.ac.uk) funded by EPSRC Tier-2 capital grant EP/T022159/1.}
}
\address{Cambridge University Engineering Dept., Trumpington St., Cambridge, CB2 1PZ U.K.\\
\small{\texttt{\{gs534,cz277,pcw\}@eng.cam.ac.uk}}}

\begin{document}
\ninept
\maketitle
\begin{abstract}
End-to-end spoken language understanding (SLU) suffers from the long-tail word problem. This paper exploits contextual biasing, a technique to improve the speech recognition of rare words, in end-to-end SLU systems. Specifically, a tree-constrained pointer generator (TCPGen), a powerful and efficient biasing model component, is studied, which leverages a slot shortlist with corresponding entities to extract biasing lists. Meanwhile, to bias the SLU model output slot distribution, a slot probability biasing (SPB) mechanism is proposed to calculate a slot distribution from TCPGen. Experiments on the SLURP dataset showed consistent SLU-F1 improvements using TCPGen and SPB, especially on unseen entities. On a new split by holding out 5 slot types for the test, TCPGen with SPB achieved zero-shot learning with an SLU-F1 score over 50\% compared to baselines which can not deal with it. In addition to slot filling, the intent classification accuracy was also improved.  
\end{abstract}
\section{Introduction}
\label{sec:intro}
Spoken language understanding (SLU) plays a key role in spoken dialogue systems, which includes user intent detection and slot-filling. SLU is often implemented as a pipeline system that first transcribes speech into text with an automatic speech recognition (ASR) system, and then performs intention detection or slot-filling with a natural language understanding (NLU) component operating only on texts. 
The pipeline systems ignore the prosody and pronunciation information embedded in the speech but not in the text and can have more NLU errors propagated from the ASR errors, in particular, named entity-related errors. 
End-to-end SLU systems \cite{e2eslu1,e2eslu2,e2eslu3} can potentially resolve these issues, by combining the ASR and NLU components into a single audio-grounded model. Such systems can be improved by leveraging powerful acoustic and language representations pre-trained with a large number of data \cite{w2v2,roberta,distill1,pretrain1, pretrain2,cti, rnntinterface, aedinterface1, w2v2slu}. 


While end-to-end SLU systems mitigate error propagation, the slot-filling task still relies on the correct recognition of the named entities in the speech, especially when the named entity contains rare words. In end-to-end ASR systems, the recognition of those valuable rare words is often addressed via contextual biasing which integrates contextual knowledge represented as a \emph{biasing list} into ASR systems\cite{shallow_context_1,shallow_context_2,shallow_context_3,deep_context_1,deep_context_2,deep_context_3,deep_context_4,deepshallow,DBRNNT,word_mapping}. The biasing list is a list of words or phrases (\emph{biasing words}) that are likely to appear in a given context. The recognition accuracy of those words can be improved if they are incorporated into the biasing list. In SLU and spoken dialogue tasks, possible named entities for each slot type can be collected to form a structured knowledge base (KB), and the biasing list can be extracted from the KB \cite{tcpgenmbr} by selecting rare and unseen entities in slots that are relevant to the current context. Therefore, applying contextual biasing in end-to-end SLU systems is both natural and beneficial.

This paper proposes integrating contextual knowledge into the end-to-end SLU system via the tree-constrained pointer generator (TCPGen) \cite{tcpgen,tcpgenmbr,tcpgengnn}. TCPGen builds a neural shortcut between the biasing list and the model output via a pointer generator mechanism and uses a prefix-tree representation to handle large biasing lists containing thousands of words. In addition to the improved recognition accuracy, the slot shortlist (SS) and slot probability biasing (SPB) methods were proposed in this paper to take full advantage of TCPGen in SLU. With SS, TCPGen in SLU can handle a more focused biasing list for better recognition accuracy by only incorporating entities from a shortlist of slot types predicted using a class language model (CLM). With SPB, TCPGen in turn provides indications of whether an entity in the biasing list has been used for recognition for SLU. This indication was done by estimating a distribution over slot types from TCPGen and adding it to the original slot-filling output via the pointer generator mechanism. In particular, this method boosts the performance of the SLU system on unseen entities, and it also enables the SLU system to achieve zero-shot learning of unseen slot types when using a biasing list of entities belonging to that type.

Experiments on the spoken language understanding resource package (SLURP) data \cite{slurp} showed consistent improvements using TCPGen in SLU, with a particular performance boost on unseen entities. Moreover, SLU systems with TCPGen and SPB achieved zero-shot learning on unseen slot types\footnote{Code available at \url{https://github.com/BriansIDP/espnet/tree/TCPGenSLU/egs/slurp/asr1}}.

The rest of this paper is organised as follows: Sec.~\ref{sec:relwork} reviews related studies. Sec.~\ref{sec:tcpgen} introduces TCPGen, followed by Sec.~\ref{sec:slu} which explains how SS and SLU can be applied with TCPGen. Sec.~\ref{sec:setup} describes the experimental setup, and Sec,~\ref{sec:result} discusses the results. Finally, conclusions are provided in Sec.~\ref{sec:conclusion}.

\section{Related Work}
\label{sec:relwork}
\subsection{Contextual biasing}
Previous studies on contextual biasing have been focused on either shallow-fusion-based score-level interpolation \cite{shallow_context_1,shallow_context_2,shallow_context_3} or deep neural representation \cite{deep_context_1,deep_context_2,deep_context_3,deep_context_4} methods. Recently, there was research using both biasing methods. In \cite{deepshallow,DBRNNT}, shallow fusion and deep biasing were applied together in the end-to-end ASR model. More recently, neural shortcuts to the final output distribution via a pointer generator \cite{tcpgen, MEM} or neural-FST \cite{neuralfst} have been proposed which can be optimised in an end-to-end fashion. TCPGen \cite{tcpgen} also achieved high efficiency by using a symbolic prefix-tree search to handle biasing lists of thousands of words. Work in \cite{tcpgengnn} used a graph neural network (GNN) to encode the prefix tree in TCPGen, which achieved further improvements in the recognition accuracy of biasing words.

\subsection{End-to-end SLU}
In recent years, end-to-end SLU systems have been trying to leverage external knowledge to achieve improved performance. Most research focused on using implicit knowledge from pre-trained representations, such as RoBERTa \cite{roberta} and wav2vec2.0 \cite{w2v2}. External knowledge from those models was integrated via knowledge distillation \cite{distill1} or network integration \cite{cti, rnntinterface, aedinterface1, w2v2slu}. Specifically, in \cite{aedinterface1} and \cite{cti}, neural representations from the ASR model and text representations from a pre-trained LM were combined in the interface for the SLU tasks. On the other hand, external contextual knowledge can also be extracted from explicitly structured KBs. Work in \cite{slucontext} applied a knowledge encoder to encode KB entries as part of the input to the NLU component. Work in \cite{tcpgenmbr} exploited the ontology in a dialogue system for improved ASR performance.

\section{Tree-constrained Pointer Generator}
\label{sec:tcpgen}

TCPGen is a neural network-based component combining the symbolic prefix-tree search with a neural pointer generator \cite{pointer_1} for contextual biasing, which enables end-to-end optimisation with ASR systems. At each output step, TCPGen calculates a distribution over all valid wordpieces constrained by a word-piece-level prefix tree built from the biasing list (referred to as the TCPGen distribution). TCPGen also predicts a generation probability indicating how much contextual biasing is needed at a specific step. The final output is the interpolation between the TCPGen distribution and the original ASR model output distribution, weighted by the generation probability.

Specifically, a set of valid wordpieces, denoted as $\mathcal{Y}^\text{tree}_i$, is obtained by searching the prefix-tree with a given history output sequence. Then, denoting $\mathbf{x}_{1:T}$ and $y_i$ as input acoustic features and output wordpieces respectively, $\mathbf{q}_i$ as the query vector carrying the history and acoustic information, and $\mathbf{K}=[...,\mathbf{k}_j,...]$ as the key vectors, scaled dot-product attention is performed between $\mathbf{q}_i$ and $\mathbf{K}$ to compute the TCPGen distribution $P^\text{ptr}$ and an output vector $\mathbf{h}^{\text{ptr}}_i$ as shown in Eqns. \eqref{eq:TCPGen_attention} and \eqref{eq:TCPGen_value}.
\vspace{-0.1cm}
\begin{equation}
    P^{\text{ptr}}(y_{i}|y_{1:i-1},\mathbf{x}_{1:T}) = \text{Softmax}(\text{Mask}(\mathbf{q}_i\mathbf{K}^\text{T}/\sqrt{d}))
    \label{eq:TCPGen_attention}
    \vspace{-0.1cm}
\end{equation}
\begin{equation}
    \mathbf{h}^{\text{ptr}}_i = \sum\nolimits_{j} P^{\text{ptr}}(y_i=j|y_{1:i-1},\mathbf{x}_{1:T})\,\mathbf{v}^\text{T}_j
    \label{eq:TCPGen_value}
    \vspace{-0.1cm}
\end{equation}
where $d$ is the size of $\mathbf{q}_i$ (see \cite{transformer}), Mask$(\cdot)$ sets the probabilities of wordpieces that are not in $\mathcal{Y}^{\text{tree}}_i$ to zero, and $\mathbf{v}_j$ is the value vector relevant to $j$. This paper specifically focuses on the attention-based encoder-decoder (AED) ASR model with TCPGen. In AED, the query combines the context vector and the previously decoded token embedding, while the keys and values are computed from the decoder wordpiece embedding, with a shared projection matrix. The generation probability which takes a value between 0 and 1, is calculated using the decoder hidden state and the TCPGen output vector $\mathbf{h}^{\text{ptr}}_i$. Then, the final output can be calculated as shown in Eqn. \eqref{eq:TCPGen_final}.
\begin{equation}
    P(y_i) = P^{\text{mdl}}(y_i)(1-{P}^\text{gen}_i) + P^{\text{ptr}}(y_i)P^\text{gen}_i
    \label{eq:TCPGen_final}
\end{equation}
where conditions, $y_{1:i-1}, \mathbf{x}_{1:T}$, are omitted for clarity. $P^{\text{mdl}}(y_i)$ represents the output distribution from the standard end-to-end model, and $P^{\text{gen}}_i$ is the generation probability. 

Instead of wordpiece embeddings, nodes on the prefix tree can also be represented with GNN encodings, as proposed in \cite{tcpgengnn}. GNN encodings provide a more powerful representation by exploiting a lookahead functionality in the tree search, where each node encodes information about future wordpieces on its branches. Specifically, a graph convolutional network (GCN) \cite{gcn} was used in this paper.

\section{TCPGen for End-to-end SLU}
\label{sec:slu}

\begin{figure}[t]
    \centering
    \includegraphics[scale=0.27]{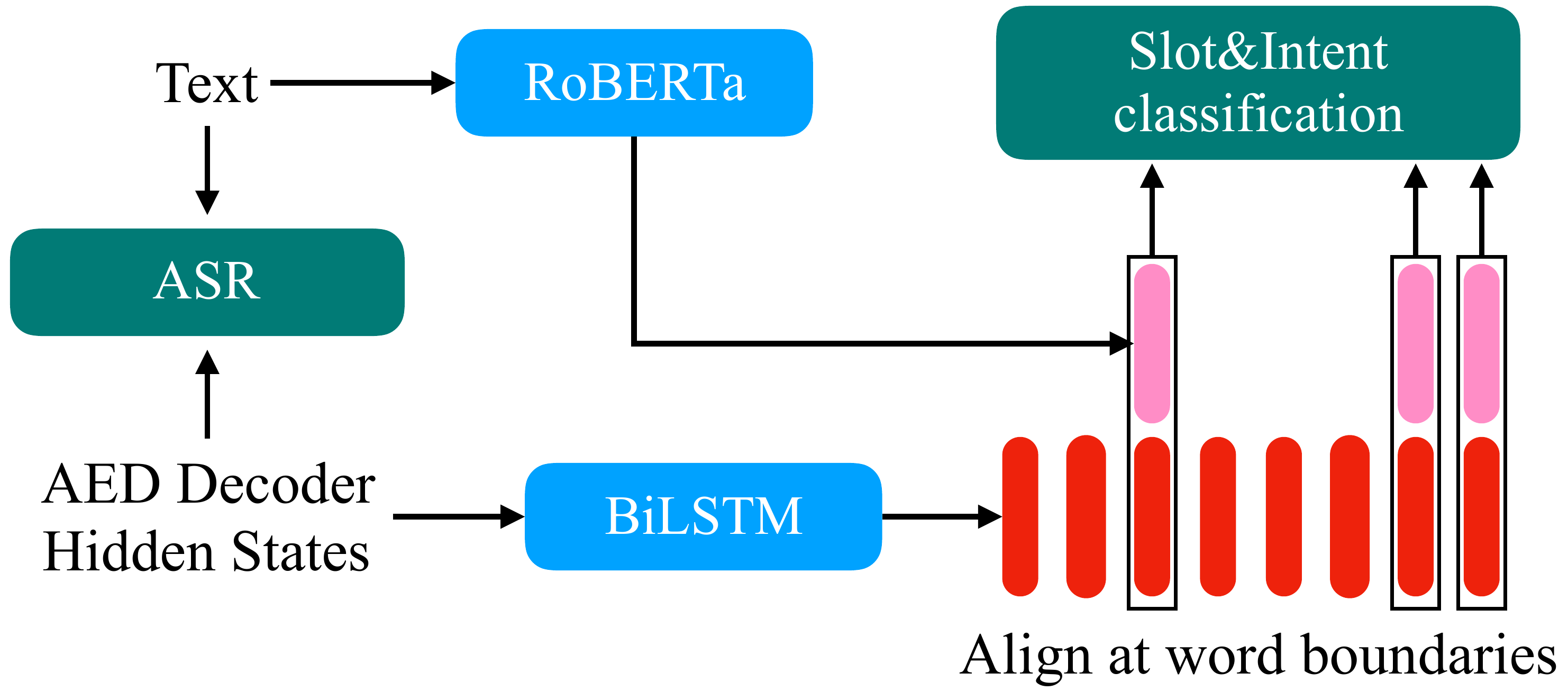}
    \vspace{-0.3cm}
    \caption{End-to-end SLU system. The word-level alignment aligns the two sequences of representations at word boundaries.}
    \vspace{-0.3cm}
    \label{fig:slu}
\end{figure}

The SLU system in this paper is shown in Fig. \ref{fig:slu}, similar to \cite{aedinterface1,rnntinterface}. The decoder hidden state sequence from AED was first sent through a bi-directional long short-term memory module. Then, to leverage external knowledge from a pre-trained representation, a sequence of RoBERTa output vectors was extracted and both sequences were aligned and concatenated at word boundaries. The concatenated vector sequence was sent to perform slot classification at the word boundary of each word. Intent classification only used the output at the final step. The AED, RoBERTa and BiSLTM modules were jointly optimised with the ASR, slot and intent classification tasks.

\subsection{TCPGen with slot shortlists}

\begin{figure}[t]
    \centering
    \includegraphics[scale=0.27]{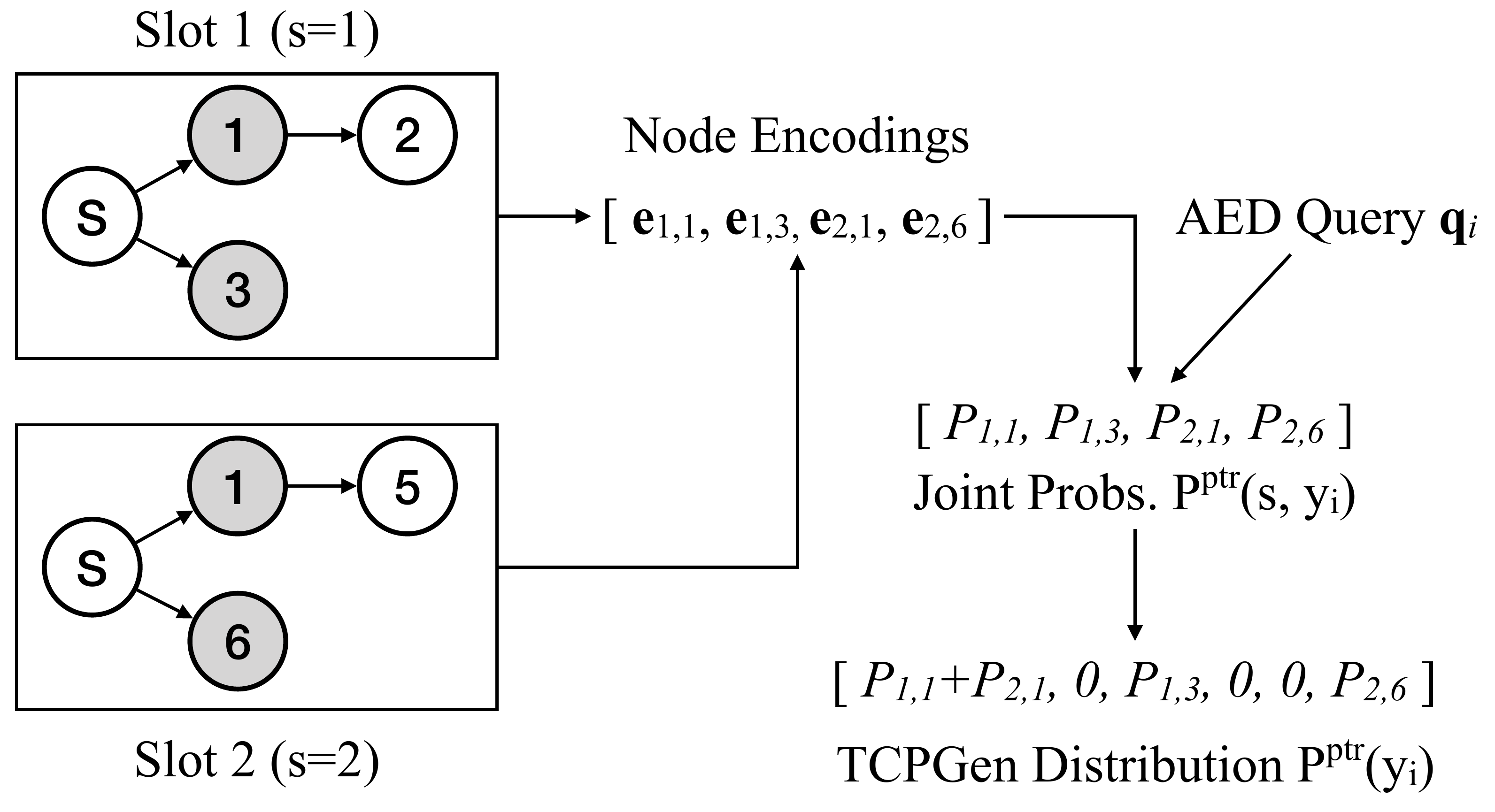}
    \vspace{-0.3cm}
    \caption{Illustration of TCPGen with SS. The example SS contains 2 slot types with their prefix trees each containing 2 entities. Nodes with grey fillings are the valid subset of wordpieces. Number 1 to 6 represents wordpieces. The current decoding step is $i$, $\mathbf{e}_{m,j}$ denotes wordpiece $j$ on tree $m$, and $P_{m,j}=P^{\text{ptr}}(s=m,y_i=j)$}
    \vspace{-0.3cm}
    \label{fig:classprob}
\end{figure}

TCPGen with slot shortlists (SS) utilised a more focused biasing list for better recognition.
Specifically, a word-level causal CLM which predicts the class of the next word given word history was applied to predict a shortlist of the top $N$ most probable slot types at the beginning of each word. The biasing list can be extracted by collecting entities belonging to those slots. However, when using GCN encodings during inference, as SS kept varying during the decoding, it was inefficient to encode the prefix tree repeatedly for every update of the biasing list. To address this, prefix trees were encoded offline for each slot type separately before decoding. The computation of the TCPGen distribution first computed the joint probability distribution over valid wordpieces and slot types and marginalised w.r.t. the slot types. This procedure is illustrated in Fig. \ref{fig:classprob}.

As shown in Fig. \ref{fig:classprob}, the number $j$ on each node denotes a wordpiece and encodings of each node, $\mathbf{e}_{m,j}$, were obtained by encoding the tree corresponding to slot $m$ using GCN. These node encodings were concatenated to form the keys and the same scaled dot-product attention in TCPGen was performed to calculate a distribution over these node encodings, as shown in Eqn. \eqref{eq:clmtcpgen}.
\begin{equation}
    P^{\text{ptr}}(s,y_i) = \text{Softmax}(\mathbf{q}^T_i[\cdots,\mathbf{e}_{m,j},\cdots]/\sqrt{d})
    \label{eq:clmtcpgen}
\end{equation}
where $\mathbf{q}_i$ and $d$ are the same as Eqn. \eqref{eq:TCPGen_attention}. Finally, probabilities of nodes corresponding to the same wordpiece were summed up to obtain the final TCPGen distribution as shown in Eqn. \eqref{eq:clmtcpgen2}
\begin{equation}
    P^{\text{ptr}}(y_i) = \sum_{s\in\mathcal{S}} P^{\text{ptr}}(s,y_i)
    \label{eq:clmtcpgen2}
\end{equation}
where $\mathcal{S}$ denotes the set of slot types.
This method is applied during inference.
Moreover, as for TCPGen, TCPGen with SS can be generalised to work with phrases. When handling a biasing list of entities of more than one word, instead of obtaining a new shortlist at each word boundary, the shortlist is only updated at word boundaries where there are no valid paths on current prefix trees. 

\subsection{TCPGen with slot probability biasing (SPB)}

A distribution over slot types can also be estimated from the joint distribution in Eqn. \eqref{eq:clmtcpgen} by summing all of the node probabilities on each slot tree, as shown in Eqn. \eqref{eq:clmtcpgen3}. 
\begin{equation}
    P^{\text{ptr}}(s) = \sum_{y_i\in \mathcal{Y}_s} P^{\text{ptr}}(s,y_i)
    \label{eq:clmtcpgen3}
\end{equation}
where $\mathcal{Y}_s$ is the set of valid wordpieces at step $i$ on the tree corresponding to slot $s$. Then, this probability was interpolated with the original SLU model output slot probabilities, weighted by the generation probability ${P}^\text{gen}_i$ indicating how likely the next wordpiece token should be taken from the prefix trees, as shown in Eqn. \eqref{eq:shortcut}.
\begin{equation}
    P(s) = P^{\text{mdl}}(s) \times (1 - \alpha {P}^\text{gen}_i) + P^{\text{ptr}}(s) \times \alpha {P}^\text{gen}_i
    \label{eq:shortcut}
\end{equation}
where $\alpha$ is the hyper-parameter between 0 and 1 to restrict the influence of TCPGen as entities found in the biasing list are not always slot values. $P^{\text{mdl}}(s)$ is the original model output slot probability. This method was particularly beneficial to entities that are unseen in the training set, and it realised zero-shot learning of unseen slots by providing a list of possible entities.

\section{Experimental Setup}
\label{sec:setup}
\subsection{Data}
Experiments were performed on the SLURP data \cite{slurp}. SLURP is a collection of 72K audio recordings of single-turn user interactions with a home assistant, annotated with scenarios, actions and entities. Experiments were first performed using the official training, validation and test split, with synthesised audio used during training following \cite{espnetslu}, to show the effectiveness of TCPGen on the standard SLU task. Additionally, a new split of the data was used by holding out utterances containing entities in five randomly selected types (\texttt{podcast\_name}, \texttt{artist\_name}, \texttt{audiobook\_name}, \texttt{business\_name}, \texttt{radio\_name}) to evaluate zero-shot learning of unseen slot types. These utterances were mixed with an equal number of randomly selected utterances with seen slots or without slots to form the test set, and the rest of the utterances were used for training and validation. Moreover, the Librispeech 960-hour read English corpus was used to pre-train the ASR part of the system before training on SLURP. For both datasets, input features used 80-dimensional (-d) log-Mel filter bank features at a 10 ms frame rate concatenated with 3-d pitch features. SpecAugment \cite{specaug} with the setting $(W,F,m_F,T,p,m_T)=(40,27,2,40,1.0,2)$, as an effective data augmentation method.

\subsection{Biasing list extraction}
The biasing list selection on Librispeech data followed \cite{tcpgengnn}. For SLURP, lists of slot entities in the data were categorised into their corresponding slots to form the KB. In this paper, rare words are defined as words in the KB and appeared less than 30 times in the SLURP training set, which also includes unseen words. There were altogether 3k rare words. The \emph{rare word biasing list} can be organised by including rare words that appeared in the list of each slot, and the \emph{rare entity biasing list} included entities containing rare words for each list. Note that unbounded slots such as date, time or frequency were not included in this biasing list as it was difficult and tedious to enumerate all possible values. The size of rare word biasing lists ranged from 4 to 712 words, and the size of rare entity biasing lists ranged from 1 to 847 entities. For experiments on the new split, all entities in the held-out 5 slots were used in the \emph{rare entity biasing list} as none of them appeared in the training set.

\subsection{Models and evaluation metrics}
The AED model used an encoder with 16 conformer blocks \cite{conformer}
with 512-d hidden state and 4-head attention, 1024-d single-head
location-sensitive attention and a 1024-d single-layer unidirectional
LSTM decoder. The BiLSTM module contained a 1024-d single-layer bi-directional LSTM. The RoBERTa base model with 768-d output representations was used. The CLM for SS prediction was a single-layer 2048d LSTM LM. AED models together with TCPGen were first trained on Librispeech 960-hour data for 20 epochs. The model parameters were then used to initialise relevant parts in the SLU system. During training, $n$ slot types were selected by finding slots that appeared in the utterance annotation and adding distracting slots to match the $n$ used during inference. Rare words belonging to those slot types were gathered to form the biasing list for training. 

Models were evaluated using word error rates (WER) and rare word error rates (R-WER) following \cite{tcpgengnn}. R-WER is the rate of deletion, substitution and insertion of a rare word. For SLU, the SLU-F1 \cite{slurp} was used to measure the slot-filling performance. The baseline is a pipeline system which takes the 1-best hypothesis from the AED model as the input to a RoBERTa-based NLU.

\section{Results}
\label{sec:result}

First, experiments with the official split of the SLURP data were performed. WER, R-WER and SLU-F1 were reported as shown in Table \ref{tab:tcpgenslu1}. The NLU part of the baseline used the same RoBERTa model structure as the end-to-end SLU system. Compared to the baseline, the end-to-end SLU system not only achieved better performance for SLU-F1 but also better performance for WER and R-WER, as slot-filling in a multi-task setup also facilitated ASR training. 

\begin{table}[t]
    \centering
    \caption{Results on the SLURP test set with the official split using TCPGen with different sizes of SS in SLU. Std. AED + NLU is the pipeline system. Full refers to using a single large biasing list containing all rare words, and top $n$ refers to using biasing lists corresponding to the top $n$ slot types. SPB was not used in this table. Entity refers to using the rare entity biasing list during inference. Improvements were statistically significant at $p\leq 0.01$.}
    \vspace{0.3cm}
    \begin{tabular}{llccc}
    \toprule
    System     &  Biasing list & WER & R-WER & SLU-F1 \\
    \midrule
     Std. AED + NLU & N/A  &  12.7\%  & 43.4\% & 77.8\% \\
     \midrule
     Std. SLU & N/A & 12.6\% & 43.0\% & 78.4\% \\
     +TCPGen & full & 12.4\% & 37.6\% & 78.9\% \\
     +TCPGen & top 1 & {12.1\%} & 37.0\% & {79.1\%} \\
     +TCPGen & top 2 & \textbf{11.9\%} & 36.2\% & \textbf{79.2\%} \\
     +TCPGen & top 5 & 12.0\% & \textbf{35.6\%} & 79.1\% \\
     +TCPGen & top 10 & 12.1\% & {36.6\%} & 79.0\% \\
     \midrule
     +TCPGen & top 2 entity & {12.0\%} & 36.8\% & \textbf{79.2\%} \\
    \bottomrule
    \end{tabular}
    \vspace{-0.3cm}
    \label{tab:tcpgenslu1}
\end{table}

Further reductions in WER, R-WER and SLU-F1 were achieved using TCPGen. When the full rare word list was used for biasing, 13\% relative R-WER reduction was achieved, which resulted in an increase of 0.5\% in SLU-F1. This increase was mainly due to the correct recognition of rare words. Then, more focused biasing lists were obtained using the predicted SS. The best number of slot types to include in the SS was found by balancing the trade-off between the size and the coverage of biasing lists. Having more slot types increased the chance to cover a specific rare word, but also increased the size of the biasing list which can degrade performance. In Table \ref{tab:tcpgenslu1}, the best WER and SLU-F1 were achieved using the top 2 slot types, which gave a 16\% relative R-WER reduction and an increase of 1.4\% in SLU-F1 compared to the baseline. Although the top 5 achieved a lower R-WER, as the size of the biasing list increased, the degradation of recognition accuracy on common words reduced the overall performance. Moreover, using rare entity biasing lists achieved a very similar performance to rare word biasing lists. 

\begin{table}[t]
    \centering
    \caption{Results on the SLURP test set with the official split using SS and SPB in SLU, with the effects of different $\alpha$ values (see Eqn. \eqref{eq:shortcut}). Std. AED + NLU is the pipeline system. The top 2 slot types were used for biasing lists. Unseen referred to the SLU-F1 score measured on entities containing out-of-training-set words. }
    \vspace{0.3cm}
    \begin{tabular}{lccc}
    \toprule
    System   & $\alpha$ & SLU-F1 (unseen) & Intent Acc.\\
    \midrule
     Std. AED + NLU & N/A  & 77.8\% (50.5\%) & 87.9\% \\
     \midrule
     Std. SLU & N/A & 78.4\% (51.1\%) & 88.6\% \\
     +TCPGen  & 0.0 & {79.2\%} (54.8\%) & 88.7\% \\
     +TCPGen & 0.5 & \textbf{79.5\%} (\textbf{57.5\%}) & \textbf{88.9\%} \\
    \bottomrule
    \end{tabular}
    \vspace{-0.3cm}
    \label{tab:tcpgenslu2}
\end{table}

The SPB method enabled the system to further exploit rare and unseen words that were correctly recognised using TCPGen but were still misclassified by the SLU output, and gave an additional performance boost as shown in Table \ref{tab:tcpgenslu2}.
By increasing $\alpha$ up to 0.5, the overall SLU-F1 increased by 0.3 compared to the best TCPGen SLU system without SPB, which led to an increase of 1.7\% in the overall SLU-F1 score compared to the baseline. The main contributor to this improvement was the recall rate. However, further increasing $\alpha$ degraded the SLU-F1, as the recall rate stopped increasing while precision degraded. Moreover, the intent accuracy was also compared here to complete the SLU task set, and although TCPGen was not intended to directly help intent accuracy, the classification accuracy was also improved by 1.0\% compared to the baseline. 

A considerable portion of the SLU-F1 improvement using TCPGen and SPB came from the improved performance on unseen words. To illustrate the advantage of SPB in TCPGen on unseen words in Table \ref{tab:tcpgenslu2}, a separate SLU-F1 was specifically calculated for unseen entities which comprised 5\% of all entities in the test set. With the best $\alpha$ value for the overall performance, an F1 score increase of 7.0\% was found on unseen entities. In contrast to the observation on the overall SLU-F1 score, the SLU-F1 score on these entities kept increasing with an increasing $\alpha$, as the recall rate on these entities had a much larger room for improvement. 

\begin{table}[t]
    \centering
    \caption{Results on the proposed held-out set with unseen slots. Std. AED + NLU is the pipeline system. TCPGen SLU in this table used biasing lists of all selected unseen slots. Unseen slots referred to the SLU-F1 on the unseen slot types only.}
    \vspace{0.3cm}
    \begin{tabular}{lcc}
    \toprule
    System   & $\alpha$ & SLU-F1 (unseen slots)\\
    \midrule
     Std. AED + NLU & N/A  & 29.7\% (0.0\%) \\
     \midrule
     Std. SLU & N/A & 30.1\% (0.0\%) \\
     +TCPGen & 0.0 & 29.6\% (0.0\%) \\
     +TCPGen & 0.5 & 42.1\% (36.6\%) \\
     +TCPGen & 1.0 & \textbf{52.0\%} (\textbf{50.2\%})\\
    \bottomrule
    \end{tabular}
    \vspace{-0.3cm}
    \label{tab:tcpgenslu3}
\end{table}

Finally, experiments were performed on the new split of SLURP for zero-shot learning, which had 5 unseen slots in the test sets. Results summarised in Table \ref{tab:tcpgenslu3}. For systems without SPB, even if the entity was correctly recognised, the model could never classify it to the slot type that was not covered by the training set. Thus, the SLU-F1 was dominated by the performance on the training set slot types for those systems. Besides, as the biasing list here only contained entities from unseen slot types, TCPGen alone was not helpful for the training set slot types, and only obtained a similar overall performance as the baseline and the standard end-to-end SLU system. 
On the other hand, SPB drastically improved SLU-F1 scores by enabling the slot-filling output to handle entities of unseen slot types. As before, using the best value found on the validation set, $\alpha=0.5$, achieved 42\% relative overall SLU-F1 improvement, which also achieved an F1 score of 36.6\% on unseen slots. Furthermore, if the proportion of unseen slots is known to be large, such as in the cross-domain application scenario, the value of $\alpha$ could be set higher. Eventually, with $\alpha=1.0$, TCPGen with SPB achieved an F1 score of 50.2\% on unseen slots. 

\vspace{-0.3cm}
\section{Conclusion}
\label{sec:conclusion}
This paper has proposed to use TCPGen in an end-to-end SLU system. TCPGen leverages a KB containing entities that were likely to appear in each slot type for contextual biasing. Besides, slot shortlists (SS) predicted by a class LM were used to obtain a more focused biasing list. Moreover, slot probability biasing (SPB) was proposed that estimates slot probabilities from TCPGen to bias the model slot prediction. Experiments on the SLURP data showed progressive improvements using TCPGen and SPB, with a large performance improvement in entities with unseen words. Intent detection accuracy was also improved with TCPGen. Furthermore, TCPGen with SPB achieved zero-shot learning, with an SLU-F1 score of 50\% on unseen slot types following the new split with held-out unseen slots in this paper, compared to a zero F1 score for systems without SPB. 


\end{document}